\documentclass[conference]{IEEEtran}
\IEEEoverridecommandlockouts
\usepackage{cite}
\usepackage[table]{xcolor}

\usepackage{amsmath,amssymb,amsfonts}
\usepackage{algorithmic}
\usepackage{graphicx}
\usepackage{textcomp}
\usepackage{wrapfig}
\usepackage{times}
\usepackage{epsfig}
\usepackage{graphicx}
\usepackage{amsmath}
\usepackage{caption} 
\usepackage{amssymb} 
\usepackage{multirow} 
\usepackage{tabularx}
\usepackage{comment}
\usepackage{utfsym}
\usepackage{amssymb}
\usepackage{graphics}
\usepackage{amsthm,amsmath,amssymb}
\newcommand{\attf}{\boldsymbol{f}_i^{att}} 
\newcommand{\sptf}{\boldsymbol{f}_i^{spa}} 

\def\BibTeX{{\rm B\kern-.05em{\sc i\kern-.025em b}\kern-.08em
    T\kern-.1667em\lower.7ex\hbox{E}\kern-.125emX}}
\begin{document}

\title{Dual Attribute-Spatial Relation Alignment for 3D Visual Grounding
}

\author{
\IEEEauthorblockN{Yue Xu}
\IEEEauthorblockA{
\textit{School of Information Science and Technology}\\
\textit{University of Science and}\\
\textit{Technology of China} \\
Hefei, China \\
xuyue502@mail.ustc.edu.cn
}
\and
\IEEEauthorblockN{Kaizhi Yang}
\IEEEauthorblockA{
\textit{School of Information Science and Technology}\\
\textit{University of Science and}\\
\textit{Technology of China} \\
Hefei, China \\
ykz0923@mail.ustc.edu.cn
}
\and
\IEEEauthorblockN{Kai Cheng}
\IEEEauthorblockA{
\textit{School of Data Science}\\
\textit{University of Science}\\
\textit{and Technology of China} \\
Hefei, China \\
chengkai21@mail.ustc.edu.cn
}
\and

\IEEEauthorblockN{Jiebo Luo}
\IEEEauthorblockA{
\textit{Department of Computer Science}\\
\textit{University of Rochester} \\
New York, United States \\
jluo@cs.rochester.edu
}
\and
\IEEEauthorblockN{Xuejin Chen}
\IEEEauthorblockA{
\textit{School of Information Science and Technology}\\
\textit{University of Science and Technology of China} \\
Hefei, China \\
xjchen99@ustc.edu.cn
}
}


\maketitle

\begin{abstract}
3D visual grounding is an emerging research area dedicated to making connections between the 3D physical world and natural language, which is crucial for achieving embodied intelligence.
In this paper, we propose DASANet, a \textbf{D}ual \textbf{A}ttribute-\textbf{S}patial relation \textbf{A}lignment \textbf{Net}work that separately models and aligns object attributes and spatial relation features between language and 3D vision modalities.
We decompose both the language and 3D point cloud input into two separate parts and design a dual-branch attention module to separately model the decomposed inputs while preserving global context in attribute-spatial feature fusion by cross attentions. 
Our DASANet achieves the highest grounding accuracy 65.1\% on the Nr3D dataset, 1.3\% higher than the best competitor.
Besides, the visualization of the two branches proves that our method is efficient and highly interpretable.

\end{abstract}


\section{Introduction}
\label{intro}

The ability to reason, describe, and locate objects in the physical world is crucial for human interaction with the environment. 
\emph{Visual grounding}, which aims to identify the visual region based on a language description, can enable computers to perform downstream tasks more effectively in many applications, like automatic driving and robot navigation.
With the development of deep learning, a series of studies have been conducted for 2D visual grounding. 
However, 3D visual grounding that aims to identify objects in 3D scenes remains a challenging problem due to more complex and diverse spatial relations between objects in 3D scenes.

The 3D visual grounding task is first investigated in Referit3D \cite{achlioptas2020referit_3d} and ScanRefer \cite{chen2020scanrefer}, while they develop the first vision-language dataset based on ScanNet \cite{dai2017scannet}. 
The linguistic descriptions of 3D visual grounding typically focus on two aspects of target objects: spatial relations and object attributes.
More emphasis is placed on spatial relations.
Specifically, 90.5\% of the descriptions in the Nr3D dataset \cite{achlioptas2020referit_3d} contain spatial prepositions, while 33.5\% describe object  attributes, like color, shape, etc.
Hence, how to reason spatial relations and object attributes, and effectively align the linguistic signals with 3D visual signals for identifying the referred object in 3D scenes is the key to 3D visual grounding.


\begin{figure}
\begin{center}
\includegraphics[width=0.96\linewidth]{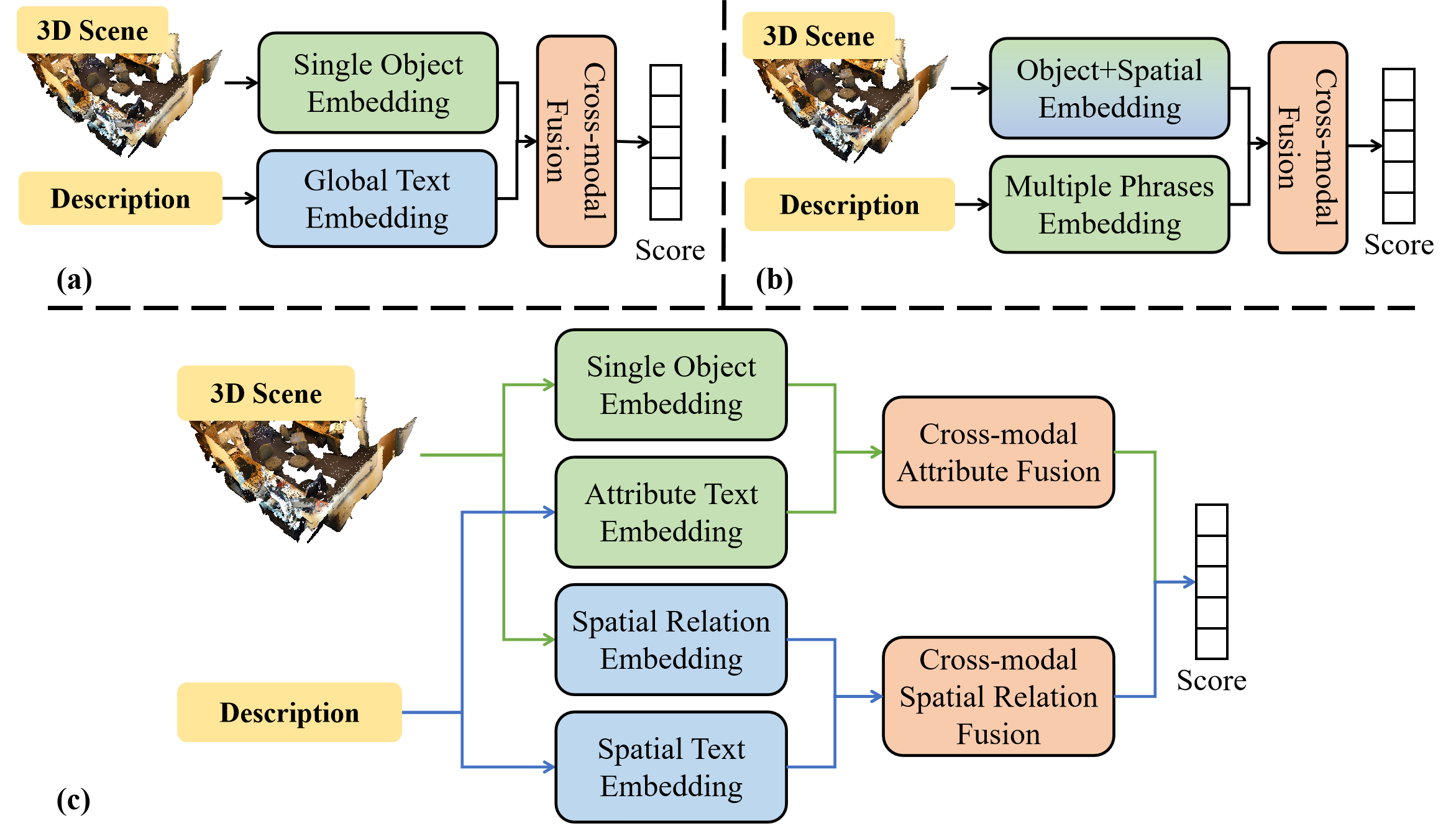}
\end{center}

\caption{Various grounding network architectures of feature embedding and cross-modal fusion in different granularity.}
\label{fig:teaser}
\vspace{-5mm}
\end{figure}

Early works \cite{transrefer3d, Huang_Lee_Chen_Liu_2021, 9710755, yuan2021instancerefer, zhao2021_3DVG_Transformer, pmlr-v164-roh22a} extract per-object visual features, then fuse the sentence-level textual feature and the object-level 3D feature to predict the grounding score, as shown in Fig.\ref{fig:teaser} (a). 
%
However, utilizing global textual information and 3D features has been proven insufficient for fine-grained cross-modal alignment, consequently resulting in ambiguous object grounding.
Recent works \cite{wu2022eda, Jain-2112} learn cross-modal alignment at finer-grained levels, as shown in Fig.\ref{fig:teaser}~(b). 
Despite the decomposition of text, the 3D objects are still represented by a single feature, leading to the entanglement of various attributes such as spatial location and semantics. 
To achieve cross-modal fine-grained alignment, two critical issues should be solved: 
(1) how to consistently align the fine-grained textual and 3D visual features; 
(2) how to exploit the global context while aligning fine-grained features to eliminate ambiguity.

In this paper, we propose the \textbf{Dual Attribute-Spatial Relation Alignment Network (DASANet)}. 
Different from other methods, our proposed DASANet explicitly decomposes the two factors, object attributes and spatial relations, in 3D visual grounding, and performs interpretable fine-grain alignment between vision-language modals.
As Fig.\ref{fig:teaser} (c) shows, both the 3D scene and the textual description are explicitly decoupled into object attributes and spatial relations.
We encode and enhance these two parts in the attribute branch and the spatial relation branch respectively, ensuring the consistency of fine-grained alignment.

To effectively exploit global context, we first apply a self-attention between two feature parts and then a cross-attention between text and visual features to incorporate the scene context information into the reasoning process.
In the end, we combine the results from both branches to obtain the final grounding score.
Besides, we propose a new training strategy using ground-truth attribute scores (GTAS) to better disentangle the attribute and spatial features while enhancing the model interpretability.

In summary, our main contributions are as follows: 
\begin{itemize}

\item[$\bullet$] We propose a novel Dual Attribute-Spatial Relation Alignment Network (DASANet) for 3D visual grounding with fine-grained visual-language alignment.

\item[$\bullet$] With our GTAS training strategy, our model exhibits better feature disentanglement and fine-grained feature alignment, while showing strong model interpretability.

\item[$\bullet$] 
Our method achieves the highest grounding accuracy on the Nr3D dataset and comparable performance on Sr3D with state-of-the-art methods.

\end{itemize}

\section{Related Work}

Referit3D \cite{achlioptas2020referit_3d} and ScanRefer \cite{chen2020scanrefer} first propose the 3D visual grounding task and construct the datasets for 3D visual grounding tasks by describing the attributes and spatial positions of 3D objects in ScanNet \cite{dai2017scannet}. 
Early methods can be divided into two categories by their network structure: Graph-based methods and Transformer-based methods.
Graph-based methods \cite{Huang_Lee_Chen_Liu_2021, 9710755, yuan2021instancerefer} model and learn spatial relations between objects in the scene based on graphs. 
Although these scene graphs explicitly represent spatial relations, it is often difficult to model long-distance spatial relations due to the graph construction mechanism based on $k$ nearest neighbors only. 
With the relation reasoning capability of Transformers, some other approaches \cite{transrefer3d, zhao2021_3DVG_Transformer, pmlr-v164-roh22a} utilize the Transformer architecture for cross-modal feature fusion. 
These methods regard 3D object features and descriptive features as tokens and feed them into the Transformer model for fusion and enhancement.
The enriched cross-modal features are employed to predict the similarity score between the language description and visual features of each object.

However, the limited scale of training data for 3D visual grounding makes the task significantly challenging than 2D visual grounding. 
Many approaches have been proposed for data augmentation or effective training with imperfect data. 
Based the Transformer-based framework, MVT \cite{huang2022multi} and ViewRefer \cite{guo2023viewrefer} introduce multi-view information of 3D scenes to eliminate the ambiguity of viewpoint and object orientation.
To alleviate the negative effect of noisy point clouds, SAT \cite{yang2021sat} incorporates 2D images into training to provide cleaner semantics. ViL3DRel \cite{chen2022vil3dref} proposes a distillation approach to facilitate cross-modal learning with teacher-student models. 
Recent works \cite{wu2022eda, Jain-2112, abdelreheem2022scanents} have shifted away from cross-modal learning solely at the object and sentence level, while focusing on feature extraction from various levels.
EDA \cite{wu2022eda} decouples the text and aligns the dense phrases with 3D objects.
ScanEnts3D \cite{abdelreheem2022scanents} provides additional annotations and losses to explore explicit correspondences between 3D objects and the description words.

Although numerous efforts have been made, existing works do not adequately disentangle the spatial relations between objects and attributes of a single object, leading to entanglement and ambiguity in these heterogeneous features.
In this work, we explicitly decouple these two types of features in both modalities, and consistently align these fine-grained features while exploiting global contexts. Our method achieves state-of-the-art performance and demonstrates strong interpretability for 3D visual grounding.

\section{Our Method}

\begin{figure*}[t]
\begin{center}
   \includegraphics[width=0.98\linewidth]{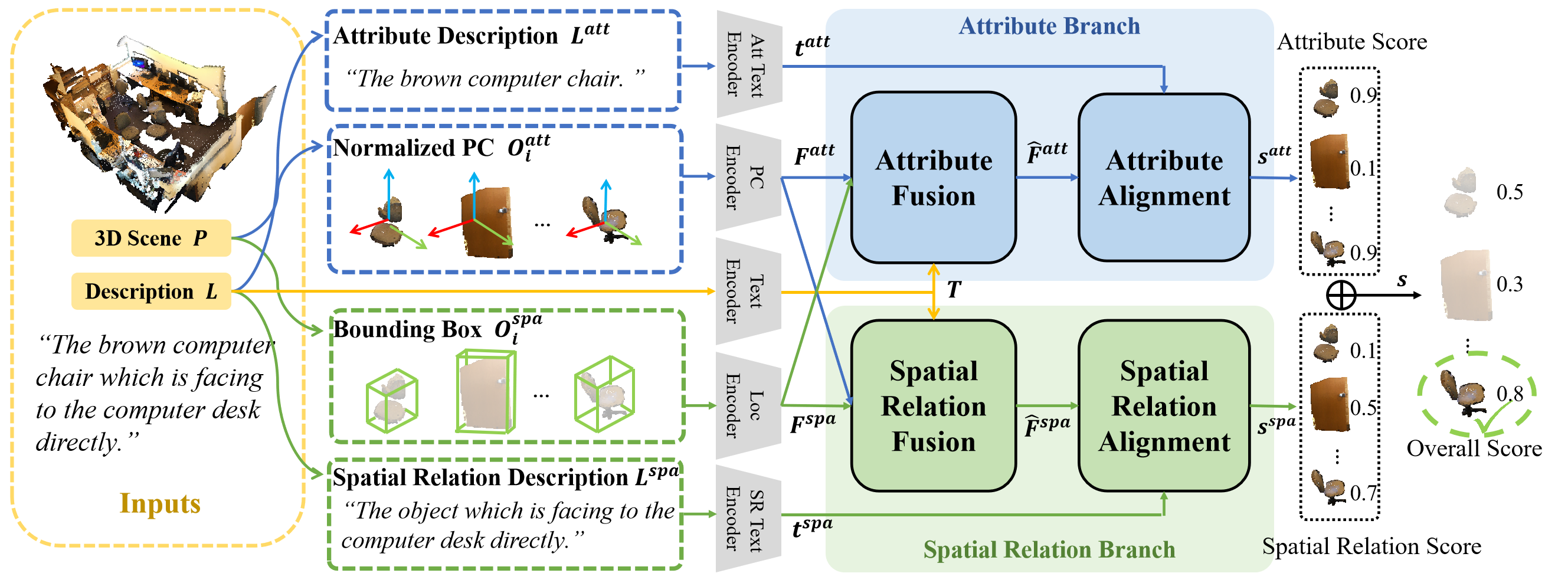}
\end{center}

   \caption{Overview of our DASANet. 
   Both the 3D point cloud and text inputs are first decomposed into the object and spatial part. Our dual-branch network, containing \emph{attribute branch} and \emph{spatial relation branch}, performs fusion and reasoning on these two aspects respectively.
   We combine the object scores of the two branches to get the final grounding results.
   }
\label{net}
\vspace{-3mm}
\end{figure*}

Given a description sentence $L$, the goal of 3D visual grounding is to localize the target object from a 3D scene represented by a 3D point cloud $P = \{p_i\}_{i=1,\ldots, N}$, $p_i=(x, y, z, r, g, b)$. 
Following the common detection-then-matching framework \cite{achlioptas2020referit_3d, chen2020scanrefer, zhao2021_3DVG_Transformer, yang2021sat}, the input point cloud is pre-segmented to $\{O_i\}_{i=1,\ldots, K}$ by ground-truth annotations. 

To align the language and 3D vision modalities, we design a Dual Attribute-Spatial Alignment Network (DASANet), a two-branch network consisting of an attribute branch and a spatial relation branch.
The text and point cloud inputs are both decomposed into the ego attribute part and spatial relation part, and
then they are fed into the attribute branch and spatial relation branch, respectively, to perform the cross-modal fusion.
In the end, we combine the per-perspective scores of the two branches to get the final grounding results.

\subsection{Decoupled Input Embedding}
\label{sec:decoupling}

\noindent\textbf{Language Description Embedding.}
We use an off-the-shelf parser \cite{honnibal-johnson-2015-improved} to extract the main subject and its adjective as attribute description $L^{att}$, and then replace them in $L$ with a common word `object' to generate the spatial relation description as $L^{spa}$. 
Using a pre-trained BERT \cite{Devlin-BERT-18}, the description $L$ with $n$ words is encoded into a token-level feature $\mathbf{T} = (\boldsymbol{t}_{cls}, \boldsymbol{t}_1, ..., \boldsymbol{t}_n) \in \boldsymbol{R}^{(n+1)\times d}$ to perform subsequent token-level cross-model fusion, while $L^{att}$ and $L^{spa}$ are encoded into sentence-level feature $\boldsymbol{t}^{att}\in \mathbb{R}^{d}$ and $\boldsymbol{t}^{spa}\in \mathbb{R}^{d}$ respectively to enable the fine-grained properties cross-modal alignment.

\noindent\textbf{3D Object Embedding.}   
For each 3D object in the scene, we normalize its point cloud $O_i^{att}\in \mathbb{R}^{N_i\times 6}$. We employ the PointNeXT \cite{qian2022pointnext} pretrained on ScanNet to encode $O_i^{att}$ into its attribute features $\boldsymbol{f}_i^{att}\in \mathbb{R}^{d}$. 
Its bounding box $O_i^{spa}=(x_c,y_c,z_c,h,w,l)$ containing the center position and the object size is embedded to $\boldsymbol{f}_i^{spa}\in \mathbb{R}^{d}$ with a linear layer.

\subsection{Dual Attribute-Spatial Fusion}

\label{sec:fusion}
 
\begin{figure}
\begin{center}
   \includegraphics[width=0.97\linewidth]{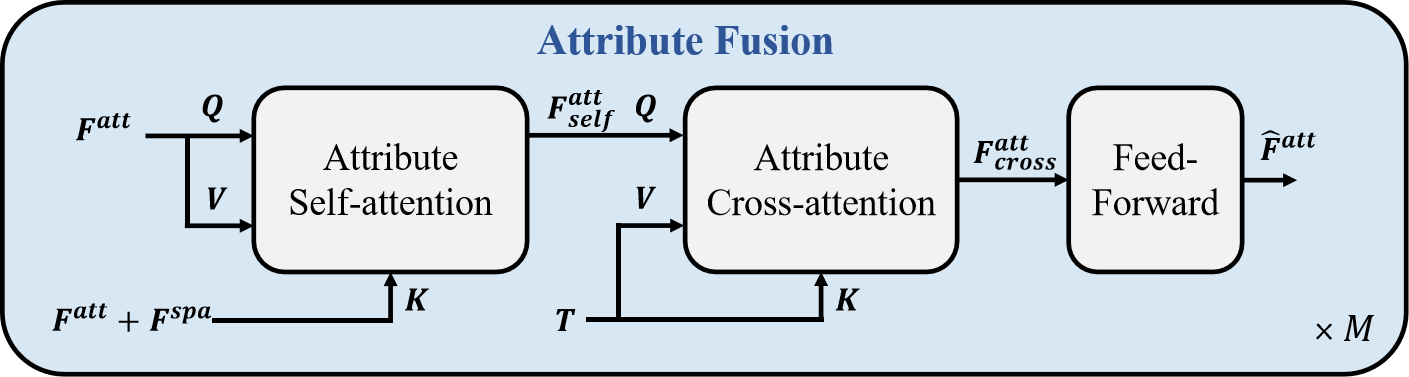}
\end{center}

   \caption{Illustration of the attribute attention module.
   }
\label{fig:attention}
\vspace{-5mm}
\end{figure}

Our DASANet consists of two symmetric branches with stacked transformer layers to separately enhance the attribute and spatial features with contextual information, as shown in Fig.~\ref{net}.
We primarily describe the network architecture and computational process of the attribute branch, as shown in Fig.~\ref{fig:attention}. The spatial relation branch shares a similar architecture with the attribute branch.

In each layer, a self-attention module is first applied to explore the correlations of all objects in the scene. 
In the attribute branch, the attribute feature $\attf$ of each object serves as the query and value embeddings.
To involve the complete object information and guarantee compatibility of the corresponding object features of the two branches, we fuse $\attf$ and $\sptf$ for a global object feature $\boldsymbol{f}_i=\boldsymbol{f}_i^{att}+\boldsymbol{f}_i^{spa}$ as the key embedding for both branches.
Taking all the object features in the scene together, $\mathbf{F}= [\boldsymbol{f}_1, \boldsymbol{f}_2, \cdots,\boldsymbol{f}_K]$ and $\mathbf{F}^{att}=[\boldsymbol{f}_1^{att}, \boldsymbol{f}_2^{att}, \cdots,\boldsymbol{f}_K^{att} ]$, we perform self-attention:
%
\begin{equation}
\mathbf{F}^{att}_{self} = softmax(\frac{( \mathbf{F}^{att}\mathbf{W}^a_q) ( \mathbf{F} \mathbf{W}^a_k)^T}{\sqrt{d}})  \mathbf{F}^{att} \mathbf{W}^a_v,
\end{equation}
where $\mathbf{W}^a_v,\mathbf{W}^a_q,\mathbf{W}^a_k$ are learnable matrices for value, query, and key embeddings. 

Then, we introduce the global text feature $\mathbf{T}$ to enhance the 3D features using a cross-attention module to incorporate context information and preserve global scene features:
\begin{equation}
\mathbf{F}^{att}_{cross}=softmax(\frac{(\mathbf{F}^{att}_{self}\mathbf{W}^c_q) (\mathbf{T}\mathbf{W}^c_k)^T}{\sqrt{d}})\mathbf{T}\mathbf{W}^c_v. 
\end{equation}

Finally, we use a feed-forward network to map $\mathbf{F}^{att}_{cross}$ to the final 3D object attribute features $\hat{\mathbf{F}}^{att}=\{\hat{\boldsymbol{f}_i}^{att}\}$. 

Similarly, the spatial relation branch uses the spatial features $\mathbf{F}^{spa}=[\boldsymbol{f}_1^{spa}, \boldsymbol{f}_2^{spa}, \cdots,\boldsymbol{f}_K^{spa}]$ as the query and value, the global object feature $\mathbf{F}$ as key in the spatial self-attention network.
Then a spatial cross-attention module and a feedforward network are respectively employed to enhance spatial features and map them to the final spatial features $\{\hat{\boldsymbol{f}_i}^{spa}\}$.


\subsection{Dual-Branch Text-3D Alignment}

\label{sec:align}
We measure the similarity between 3D objects and textual descriptions in terms of attributes and spatial relations separately.
Following \cite{wu2022eda}, we adopt the CLIP-like \cite{Radford-21} manner to compute the similarities.
Specifically, we obtain the attribute score $s_i^{att}$ for an object $O_i$ by calculating the cosine similarity between its attribute feature $\hat{\boldsymbol{f}_i}^{att}$ and the textual description embedding of attribute $\boldsymbol{t}^{att}$, which can be formulated as:
\begin{equation}
    s_i^{att} = Sim_{cosine}(\hat{\boldsymbol{f}_i}^{att}\textbf{W}_o, \boldsymbol{t}^{att}\textbf{W}_t),
\end{equation}
where $\textbf{W}_o,\textbf{W}_t \in \mathbb{R}^{d\times d}$ are learnable matrices. 
The spatial relation score $s_i^{spa}$ is computed from $\hat{\boldsymbol{f}_i}^{spa}$ and $\boldsymbol{t}^{spa}$. 

Finally, by integrating the similarity scores $s_i^{att}$ and $s_i^{spa}$ from both branches, we obtain the overall matching score $s_i$ for the object $O_i$ with the description $L$:
\begin{equation}
s_i = s_i^{att} + s_i^{spa}.
\end{equation}

\begin{table*}[t]
\centering

\caption{Grounding accuracy comparison on Nr3D and Sr3D datasets. We color each cell as \colorbox{red!40}{best} and \colorbox{orange!40}{second-best}. The `hard' data contains more than two distractors in the scene, while the others are `easy'. VD. indicates `view-dependent' data, which requires the observers to face certain directions, while the VI. (view-independent) data does not.}
\label{table1}
\begin{tabular}{c c| c c c c c|c c c c c}
\hline
& &&&Nr3D&&&&&Sr3D\\

 Method& Venue & Overall& Easy& Hard& VD.& VI. & Overall& Easy& Hard& VD.& VI.\\
\hline
ReferIt3DNet \cite{achlioptas2020referit_3d} &ECCV2020 & 35.6 $\pm$0.7&43.6&27.9& 32.5&37.1 &40.8$\pm$0.2& 44.7 &31.5 &39.2 &40.8\\
TGNN \cite{Huang_Lee_Chen_Liu_2021}& AAAI2021&37.3$\pm$0.3 &44.2& 30.6 &35.8 &38.0&45.0$\pm$0.2&48.5&36.9&45.8&45.0\\
IntanceRefer \cite{yuan2021instancerefer}&ICCV2021 &38.8$\pm$0.4&46.0&31.8&34.5&41.9&48.0$\pm$0.3&  51.1 &40.5& 45.4& 48.1\\
3DVG \cite{zhao2021_3DVG_Transformer}&ICCV2021&40.8$\pm$0.2&48.5&34.8&34.8&43.7&51.4$\pm$0.1&54.2&44.9&44.6&51.7\\
TransRefer3D \cite{transrefer3d}&MM2021 & 42.1$\pm$0.2 &48.5 &36.0 &36.5& 44.9& 57.4$\pm$0.2& 60.5& 50.2 &49.9 &57.7\\
LanguageRefer \cite{pmlr-v164-roh22a} &CoRL2021&43.9 & 51.0 & 36.6& 41.7 & 45.0  &56.0 & 58.9 & 49.3& 49.2 & 56.3 \\
SAT \cite{yang2021sat} &ICCV2021& 49.2$\pm$0.3& 56.3& 42.4& 46.9& 50.4&57.9 $\pm$0.1&61.2 &50.0 &49.2& 58.3\\
MVT \cite{huang2022multi}&CVPR2022& 55.1$\pm$0.3 &61.3 &49.1 &54.3& 55.4&64.5$\pm$0.1& 66.9 &58.8& 58.4 &64.7 \\
ViL3DRel \cite{chen2022vil3dref}&CVPR2022 & \cellcolor{orange!40}63.8$\pm$0.5 & \cellcolor{orange!40}70.3 & \cellcolor{orange!40}57.5 & \cellcolor{orange!40}61.7 & \cellcolor{orange!40}64.8 & \cellcolor{red!40}72.8$\pm$0.2 & \cellcolor{red!40}74.9& \cellcolor{orange!40}67.9 & \cellcolor{orange!40}63.8 & \cellcolor{red!40}73.2 \\
EDA \cite{wu2022eda}&CVPR2023&52.1&58.5&46.1&50.2&53.1& 68.1& 70.3 & 62.9 &54.1& 68.7\\
ViewRefer \cite{guo2023viewrefer}&ICCV 2023 & 56.0 & 63.0 & 49.7 & 55.1 & 56.8 & 67.0&68.9&62.1&52.2&67.7\\
Our DASANet &-& \cellcolor{red!40}65.1$\pm$0.2& \cellcolor{red!40}72.9& \cellcolor{red!40}57.8& \cellcolor{red!40}62.6& \cellcolor{red!40}66.2& \cellcolor{orange!40}72.7$\pm$0.1 & \cellcolor{orange!40}74.6 & \cellcolor{red!40}68.0 & \cellcolor{red!40}65.5 & \cellcolor{orange!40}73.0 \\
\hline

\end{tabular}
\label{table_MAP}

\end{table*}

\vspace{-3pt}
\subsection{Optimization and Training Strategy}
\vspace{-3pt}
\label{sec:opt}

Following previous works \cite{achlioptas2020referit_3d, transrefer3d, huang2022multi, yang2021sat}, we use the grounding prediction loss $\mathcal{L}_{ref}$, object classification loss $\mathcal{L}_{fg}$, and text classification loss $\mathcal{L}_{text}$ to form the main loss $\mathcal{L}_{main}=\mathcal{L}_{ref}+\mathcal{L}_{fg}+\mathcal{L}_{text}.$
We adopt the distractor loss $\mathcal{L}_{dis}$, anchor prediction loss $\mathcal{L}_{anc}$ and cross-attention map loss $\mathcal{L}_{attn}$ in \cite{abdelreheem2022scanents} to form the auxiliary loss $\mathcal{L}_{aux}=\mathcal{L}_{dis}+\mathcal{L}_{attn}+\lambda\mathcal{L}_{anc}$, where $\lambda=10$. 
We adopt the teacher-student training strategy in ViL3DRel \cite{chen2022vil3dref}. We use the GT object class and colors as the attribute branch's input to train the teacher model, and distill the knowledge to the student model which takes point clouds as input.
For the teacher model training, the overall loss function can be formulated as $\mathcal{L}_{teacher}=\mathcal{L}_{main}+\mathcal{L}_{aux}$. The overall loss for the student model training is $\mathcal{L}_{student}=\mathcal{L}_{main}+\mathcal{L}_{distill}$.
Please refer to the previous works \cite{achlioptas2020referit_3d, chen2022vil3dref, abdelreheem2022scanents} for more detailed explanations of the losses.

Different from \cite{chen2022vil3dref} which only outputs a single grounding score for each object, our dual-branch network outputs the attribute score and spatial score separately. 
To better disentangle the attribute and spatial features, we design a Ground-Truth Attribute Scores (GTAS) training strategy.
To train the spatial branch, we replace the predicted attribute scores $s_i^{att}$ with the GT attribute score $g_i^{att}$, which is defined as follows:
\begin{equation}
   g_i^{att}=\begin{cases}
    1,  & C_{gt}(O_i)=C_{gt}(O_{target});\\
    -1,  &else,
    \end{cases}
\end{equation}
where $C_{gt}$ is the ground-truth category of an object. 
By equalizing the scores of the same category objects in the attribute branch, the network is enforced to distinguish the target object from distractors through spatial relationship reasoning.
While the training with GTAS strategy aids the spatial branch in learning discriminative features, the attribute branch is not fully optimized, as it only utilizes the GT attribute score instead of the predicted one.
Therefore, we introduce a fine-tuning stage after the GTAS spatial training stage, which incorporates both predicted attribute scores and spatial scores to obtain the final score for training.
The GTAS spatial training stage and fine-tuning stage are employed in both the training of the teacher and student models.

\begin{figure*}[t]
\begin{center}
   \includegraphics[width=0.97\linewidth]{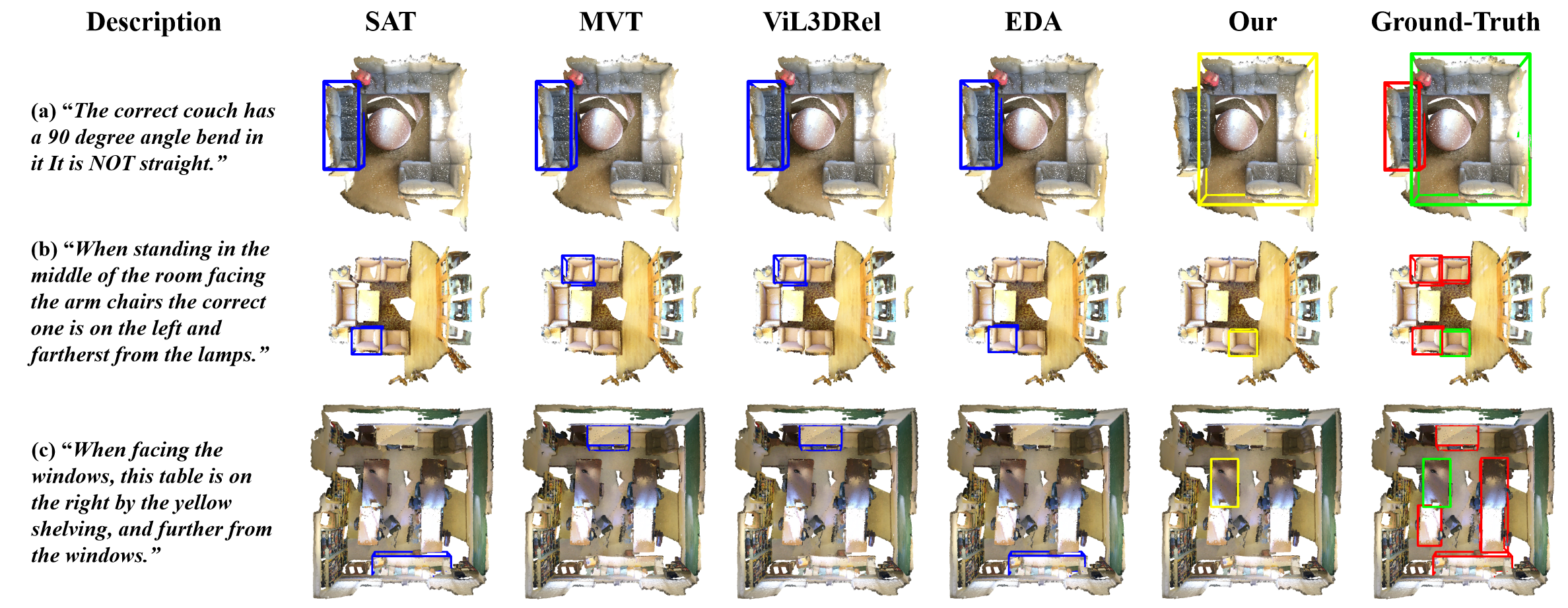}
\end{center}

\caption{Qualitative comparison of the grounding results in the Nr3D dataset. 
Our grounding results are highlighted with yellow boxes, and the results from other methods are presented with blue boxes. In the ground truth, green boxes represent the target objects, while red boxes denote distractors (objects of the same category as the target).}
\label{fig: campare}
\vspace{-3mm}
\end{figure*}

\section{Experiments and Analysis}

\noindent\textbf{Dataset and metrics.}
Our model is evaluated on the Nr3D and Sr3D datasets, consisting of descriptions for objects in the 3D indoor scene point cloud dataset ScanNet \cite{dai2017scannet}.  
There are 37, 842 human-written sentences in Nr3D, and 83, 572 automatically synthesized sentences in Sr3D. 
Nr3D dataset as a human natural language dataset, is the main dataset for 3D visual gounding research. It provides rich natural language descriptions that reflect people's descriptive habits and ways of understanding 3D scenes. In contrast, the Sr3D dataset, as a textual dataset generated by a simple template,  is intended to better assist the learning of 3D visual grounding tasks.

The evaluation metric is the accuracy of selecting the correct target among all proposals in the scene, following the default setting in ReferIt3D \cite{achlioptas2020referit_3d}.

\noindent\textbf{Implementation details.}
We employ a pre-trained BERT as our text encoder and fine-tune it during training. 
For the point clouds, we employ the PointNeXT \cite{qian2022pointnext} model pre-trained for object classification on ScanNet as the encoder and freeze it during training.
For our DASANet, we stack $M=4$ transformer layers to capture higher-order correlations. 
The hidden layer dimension $d=768$ and the number of attention heads $h=12$.
The batch size is 128 during training. 
For the Nr3D dataset, the teacher model is trained for 50 epochs in the GTAS spatial training stage and 20 epochs in the fine-tuning stage. The student is trained for 20 and 10 epochs in the GTAS spatial training and fine-tuning stage. 
For the Sr3D dataset, the training process consists of 25, 10, 10, and 10 epochs respectively in these teacher-student  training stages.

\subsection{Comparison to State-of-the-Art}

We compare our 3D visual grounding performance with existing works quantitatively and qualitatively.
In Table \ref{table_MAP}, we report the accuracy comparison of our DASANet with other methods on the Nr3D and Sr3D datasets.
Benefiting from its powerful spatial reasoning capability, our DASANet achieves the highest accuracy on Nr3D with 65.1\%, which is 1.3\% higher than the second-best ViL3DRel.
Our method also achieves comparable overall performance with the state-of-the-art method on the Sr3D dataset, and performs the best on the hard and VD. data, particularly outperforming the second-best method by 1.7\% on the VD. data.
Note that the Nr3D dataset is constructed with manual descriptions and contains much more free-form texts, which is more challenging than the template-based Sr3D dataset.
In addition, our method exhibits high stability and robustness to random seeds, as evidenced by the low standard deviation in accuracy under five different random seeds (0.2 for Nr3D and 0.1 for Sr3D).

\begin{figure}[t]
\begin{center}
   \includegraphics[width=0.98\linewidth]{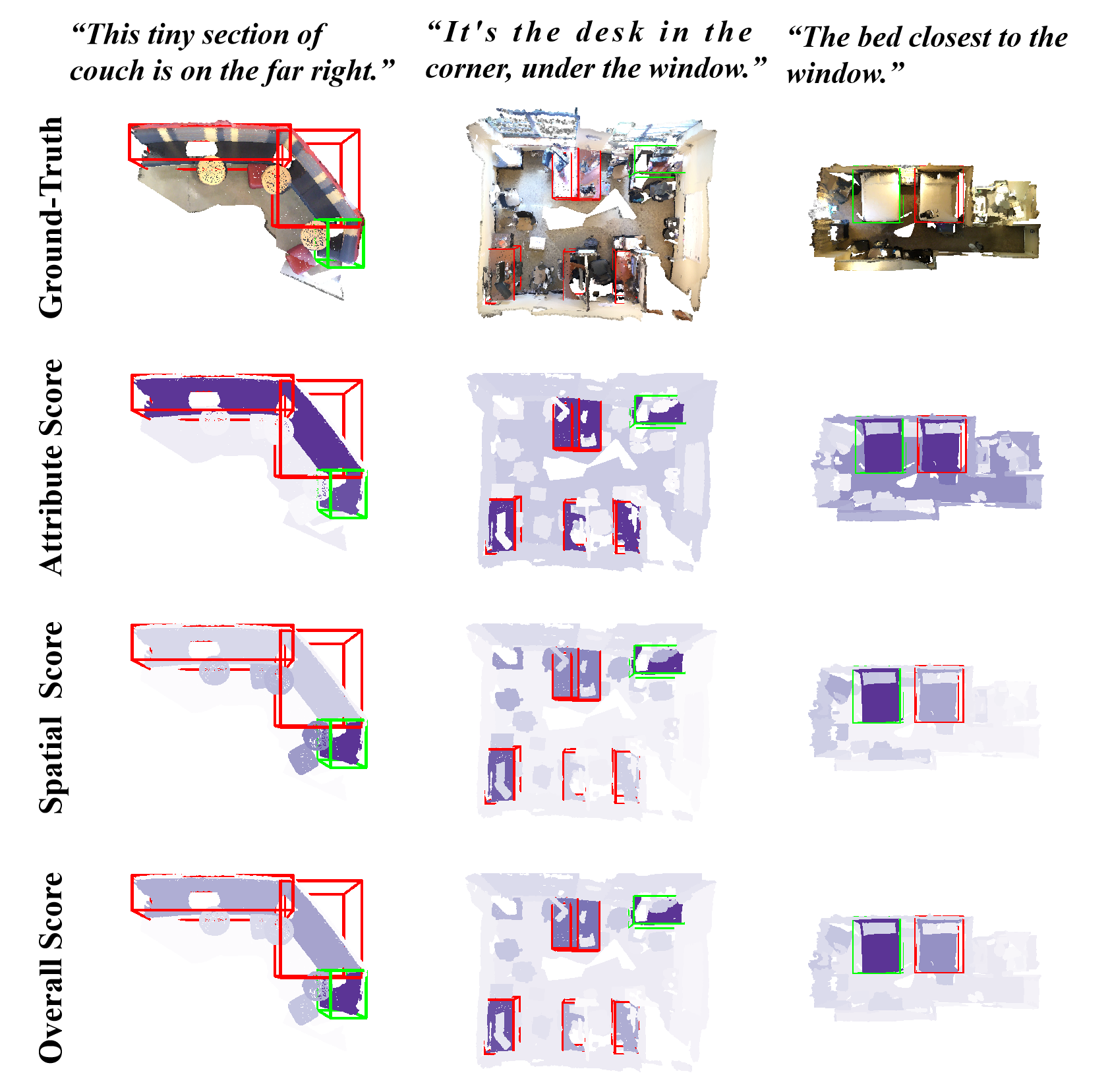}
\end{center}

\caption{Visualization of the attribute, spatial relation, and overall scores in our dual-branch network.
}
\label{score}
\vspace{-4mm}
\end{figure}

Table \ref{table:scanrefer} presents results on ScanRefer dataset with ground-truth object proposals. DASANet achieves 62.3\% on the ScanRefer dataset with a standard deviation of 0.1\%, which is 2.4\% better than the second-best method, ViL3DRel.
\begin{wraptable}{r}{4.7cm}
    \centering
    \caption{ Grounding accuracy
(\%) on ScanRefer with ground-truth object proposals.
}
    \begin{tabular}{c c}
    \hline
    Method &Acc\\
    \hline
    ScanRefer\cite{chen2020scanrefer} & 44.5\\
    ReferIt3DNet\cite{achlioptas2020referit_3d} &46.9$\pm$0.2\\
    SAT\cite{yang2021sat} &53.8$\pm$0.2\\
    MVT\cite{huang2022multi} &54.8$\pm$0.1\\
    ViL3DRel\cite{chen2022vil3dref} & 59.9$\pm$0.2\\
    Our DASANet &\textbf{62.3$ \pm$0.1}\\
    \hline
    \end{tabular}
    \label{table:scanrefer}
\end{wraptable}
The text description in the ScanRefer dataset consists of two separate sentences for attribute description and spatial relation description, respectively.
With this data form, our method takes full advantage of dual-branch fine-grained alignment of attributes and spatial relations, which further proves DASANet's superiority.

We show the qualitative comparison with \cite{wu2022eda, huang2022multi, yang2021sat, chen2022vil3dref} in Fig.\ref{fig: campare}.
As Fig. \ref{fig: campare} (a) shows, despite employing a dual-branch network, our method is capable of accurately grounding with text that only consists of complex attribute descriptions.
Fig. \ref{fig: campare} (b,c) shows two examples of complex descriptions containing multiple objects and complicated relations, where our method correctly predicts the target object, while other methods fail.
This demonstrates the effectiveness and reliability of our method in handling more natural and lengthy language descriptions.

\subsection{Interpretability Analysis}

Benefiting from the fine-grained decoupling and alignment, DASANet exhibits high interpretability.
We align text and 3D objects in terms of the object attribute and spatial relations.
We use the attribute score $s_i^{att}$ and spatial relation score $s_i^{spa}$ to respectively represent the similarity between 3D objects and text in corresponding aspects.
Though only the final scores $s_i$ is used for supervision, the inetermediate $s_i^{att}$ and $s_i^{spa}$ demonstrate the interpretability.

Fig.~\ref{score} shows three examples of $s_i^{att}$, $s_i^{spa}$, and $s_i$, where darker colors indicate higher scores.
The attribute branch predicts similar high scores for objects that align with the attribute description, for example, the three `couches', six `desks', and two `beds' in Fig.~\ref{score}.
The spatial relation branch distinguishes objects with similar attributes through spatial relation reasoning and predicts distinct scores for the target object and distractors. 
In the first two examples, the objects `on the far right' and `closest to the window' are obviously darker than other objects.
In the last example, the sentence describes two spatial relations, `in the corner' and `under the window'. 
The objects that only satisfy one spatial relation are slightly darker, while the target that fulfills both is the darkest.
Combining the two branches, our DASANet can predict the target object with strong interpretability.

\subsection{Ablation Study}
\label{Sec: ablation}

\begin{table}[!tb]
\centering

\caption{Ablation studies of our method on the Nr3D dataset.}
\begin{tabular}{c c c c c }
\hline
&Att-Spa fusion&GTAS& Overall\\
\hline
$R_1$ & concat
& \scalebox{0.85}[1]{$\times$}&62.6
\\
$R_2$ & add
& \scalebox{0.85}[1]{$\times$}&63.2
\\

\hline
$R_3$ &  concat
& \raisebox{.6ex}{\scalebox{0.7}{$\sqrt{}$}}
&64.4\\
$R_4$ & add
& \raisebox{.6ex}{\scalebox{0.7}{$\sqrt{}$}}
&\textbf{65.1}\\
\hline

\end{tabular}
\label{ablation}
\vspace{-2mm}
\end{table}

We conduct ablation experiments on the feature fusion manner used in the self-attention module and the proposed GTAS training strategy to validate their effectiveness.



\noindent\textbf{Att-Spa. feature fusion.}
In our dual-branch feature fusion and alignment pipeline, we fuse the separate object attribute feature $\boldsymbol{f}_i^{att}$ and the spatial relation feature $\boldsymbol{f}_i^{spa}$ as a global object featrue $\boldsymbol{f}_i$ for the key embedding in our intra-modal attention module.
Here we explore different attribute-spatial (Att-Spa) feature fusion manner, including concatenation ($R_1$ and $R_3$) and summation ($R_2$ and $R_4$).
The experiment results shown in Table~\ref{ablation} indicate that directly adding the two features for fusion yields higher performance ($R_2$ and $R_4$), which is the way we use in our method.

\noindent\textbf{Training with GT attribute scores.}
We validate the effectiveness of the proposed ground-truth attribute scores (GTAS) training strategy described in Sec.~\ref{sec:opt}. 
As shown in Table \ref{ablation}, $R_3$ and $R_4$ with GTAS training strategy achieves 64.4\% and 65.1\%, higher than $R_1$ and $R_2$. 
It demonstrates that with our decoupling-based framework, the GTAS training strategy fully exploits the independence of the two distinct properties and effectively improves grounding performance.

\section{Conclusion}

In this paper, we propose a dual-branch grounding network to learn spatial reasoning more effectively for the 3D visual grounding task. 
While incomplete and noisy point clouds in complex scenes bring significant challenges in identifying 3D objects, we design decoupled embedding and alignment for single-object attributes and inter-object spatial relations to disentangle these two heterogeneous features. 
We decompose the 3D visual grounding task into two sub-tasks, cross-modal object attribute alignment and spatial relation alignment between the language description and point cloud input.  
Based on the dual-branch network architecture, we propose a novel training strategy that uses ground-truth attribute scores first to enforce the network to learn more discriminative spatial relationship features from the imperfect point clouds. 
Our DASANet achieves new state-of-the-art prediction accuracy with high interpretability in spatial reasoning.
We also found that data deficiency is a critical factor limiting the development of 3D grounding technology. 
Data augmentation with large language models and 3D visual content generation approaches is worth studying in the future. 


\end{document}